\pdfoutput=1

\documentclass[11pt]{article}
\usepackage[preprint]{acl}

\usepackage{times}
\usepackage{latexsym}

\usepackage[T1]{fontenc}

\usepackage[utf8]{inputenc}

\usepackage{microtype}

\usepackage{inconsolata}

\usepackage{graphicx}

\newif\ifwithappendix
\withappendixfalse

\newif\ifappendixshown
\appendixshowntrue

\usepackage[T1]{fontenc}
\usepackage[utf8]{inputenc}
\usepackage[textsize=tiny]{todonotes}
\usepackage{graphicx}
\usepackage{booktabs}
\usepackage{latexsym}
\usepackage{microtype}

\usepackage{footnote}
\makesavenoteenv{tabular}
\makesavenoteenv{table}
\makesavenoteenv{table*}

\newcommand\minput[1]{%
  \input{#1}%
  \ifhmode\ifnum\lastnodetype=11 \unskip\fi\fi}

\usepackage{hyperref}
\usepackage{xstring}

\newcommand{\noqa}[1]{}
\newcommand{\noqall}[1]{}

\usepackage{amsmath}
\usepackage{amssymb}
\usepackage{multirow}
\usepackage{algorithm}
\usepackage[noend]{algpseudocode}
\usepackage[capitalise]{cleveref}
\crefname{algorithm}{Algorithm}{Algorithms}
\crefname{figure}{Figure}{Figures}
\usepackage{comment}
\usepackage{graphicx}
\usepackage{float}

\title{Tackling prediction tasks in relational databases with LLMs}

\author{Marek Wydmuch$^{\ast,\dag}$ \\
  \And
  Łukasz Borchmann$^{\ast}$ \\
  $^{\ast}$ Snowflake AI Research \\
  \{first-name\}.\{last-name\}@snowflake.com \\
  $^{\dag}$ Poznan University of Technology / Poznań, Poland \\
  $^{\ddag}$ Adam Mickiewicz University / Poznań, Poland \\
  \And
  Filip Graliński$^{\ast,\ddag}$ \\
  }

\begin{document}
\maketitle

\newcommand{\new}[1]{{\color{orange} #1}}

\begin{abstract}

Though large language models (LLMs) have demonstrated exceptional performance across numerous problems, their application to predictive tasks in relational databases remains largely unexplored.
In this work, we address the notion that LLMs cannot yield satisfactory results on relational databases due to their interconnected tables, complex relationships, and heterogeneous data types.
Using the recently introduced RelBench benchmark, we demonstrate that even a straightforward application of LLMs achieves competitive performance on these tasks. 
These findings establish LLMs as a promising new baseline for ML on relational databases and encourage further research in this direction.
\end{abstract}

\section{Introduction}

The application of large language models and, in general, foundational models to relational databases remains largely uncharted territory.
While the number of resources for tabular data and structured data is recently rapidly growing, with
resources available for both unsupervised and supervised learning tasks (e.g., GitTables \citep{Hulsebos_2023}, TabLib \citep{eggert2023tablibdataset627mtables}), and evaluation datasets (see the list by \citet{gardner2024large}) -- relational databases have not received
comparable attention, particularly in the context of LLMs.
\looseness=-1

Relational databases are inherently more complex than single tables due to their collections of interconnected tables linked by primary and foreign keys, encompassing one-to-many and many-to-many relationships and heterogeneous data types. The number of machine learning benchmarks that involve such databases is limited. Examples mostly include Text-to-SQL datasets such as Spider~\citep{yu-etal-2018-spider}, Bird~\citep{li2024can}, and only a few notable benchmarks focusing on prediction tasks have been published so far:
CTU Prague Relational Learning Repository~\citep{motl2024ctupraguerelationallearning}, SJTUTable~\citep{li2024rllm}, and RelBench~\citep{robinson2024relbench}. %
\looseness=-1

In this work, we focus on RelBench, a recently published, realistic, and readily available benchmark for the classification and regression on relational data. 
It is a collection of 7 relational databases from different domains, each with a set of predictive tasks, giving a total of 30 tasks. 
The challenge of the RelBench benchmark comes from the nature of relational data.
Before applying classical machine learning models, one needs to `flatten' the complex relational structure into a single table that will serve as a representation of instances.
Previous works on RelBench report results using standard machine learning approaches (e.g., gradient-boosted trees~\citep{friedman2001greedy}) and the relational deep learning method \cite{fey2023relational,robinson2024relbench}.
To date, no results using LLMs--trained in the standard manner or with tabular data--have been reported on it.
In this paper, we present a series of results obtained using pre-trained LLMs on RelBench. Importantly, we demonstrate that by traversing links between tables, one can construct information-rich documents that allow LLMs to make predictions competitive to the relational deep learning approach. Our results establish a new simple baseline and motivate further research in this direction.

\section{Relational data and RelBench}

\newcommand{\bx}{\boldsymbol{x}}
\newcommand{\by}{\boldsymbol{y}}
\newcommand{\calT}{\mathcal{T}}
\newcommand{\calL}{\mathcal{L}}
\newcommand{\calD}{\mathcal{D}}
\newcommand{\Tables}{\calT}
\newcommand{\Links}{\calL}
\newcommand{\TrainDataset}{\calD_{\mathrm{train}}}
\newcommand{\Table}{T}
\newcommand{\Link}{L}
\newcommand{\Row}{R}
\newcommand{\TableFKey}{\Table_{\mathrm{fkey}}}
\newcommand{\TablePKey}{\Table_{\mathrm{pkey}}}
\newcommand{\TableTrain}{\Table_{\mathrm{train}}}
\newcommand{\TableVal}{\Table_{\mathrm{val}}}
\newcommand{\TableTest}{\Table_{\mathrm{test}}}

A table $\Table$ is defined as a collection of $n$ rows $\Table = \{\Row, \dots, \Row_n\}$.
All the rows in the same table have the same set of attributes, also called columns.
Following the nomenclature used by \citet{robinson2024relbench}, we also refer to a single row from a single table as an entity.
With each column, we associate a data type, such as numerical, time, text, image, etc.
A relational database can be represented as a collection of tables $\Tables = \{\Table_1, \dots \Table_t\}$
and links between those tables $\Links = \{\Link_1, \dots, \Link_l\}$.
A single link $L = (\TableFKey, \TablePKey)$ consists of a foreign key column in $\TableFKey$ that points to a primary key column of $\TablePKey$.
The primary key is a unique identifier of a row in a table, while the foreign key is a reference to a primary key in another table.
Furthermore, each row in a table can be associated with a timestamp $t$, which indicates the time the row was created.
Tables that have a timestamp column are called fact tables,
while tables that do not have a timestamp and contain static information are called dimension tables \citep{garcia-molina2008databases}.

In RelBench, each predictive task is defined by a task table, which contains foreign keys to the main tables in the database, the
prediction time (also called the seed time) and the target variable $y$ to be predicted.
The task table is divided into train, validation, and test tables based on the prediction time $t_p$ (the same entities may appear multiple times in train, validation, and test tables with different timestamps).
The prediction time indicates at which time the target is to be predicted.
It filters future data in the database to ensure the temporal consistency of the data and prevent information leaks.
Formally, when predicting for an entity with prediction time $t_p$, the model can get as input only information from another entity $v$ with timestamp $t_v < t_p$.

RelBench provides 3 types of predictive tasks: (1) entity-wise binary classification tasks where $y \in \{0, 1\}$ (e.g., churn prediction), evaluated using area under the ROC curve (AUROC)~\citep{mcneil1984statistical}, (2) entity-wise regression tasks where $y \in \mathbb{R}$ (e.g., prediction of the total amount of sold items), evaluated using mean absolute error (MAE), and (3) tasks predicting a link between two entities (e.g., user item recommendation), evaluated using mean average precision (MAP).
In summary, RelBench aims to represent typical prediction and forecasting tasks in real-world relational databases.

\section{Straightforward application of LLMs}

\newcommand{\NInContext}{n_{\mathrm{inc}}}
\newcommand{\NRelated}{n_{\mathrm{rel}}}
\newcommand{\NNested}{n_{\mathrm{nest}}}
\newcommand{\NTrain}{n_{\mathrm{train}}}
\newcommand{\Depth}{d}

\begin{figure*}[ht]
\centering
\includegraphics[width=1.0\linewidth]{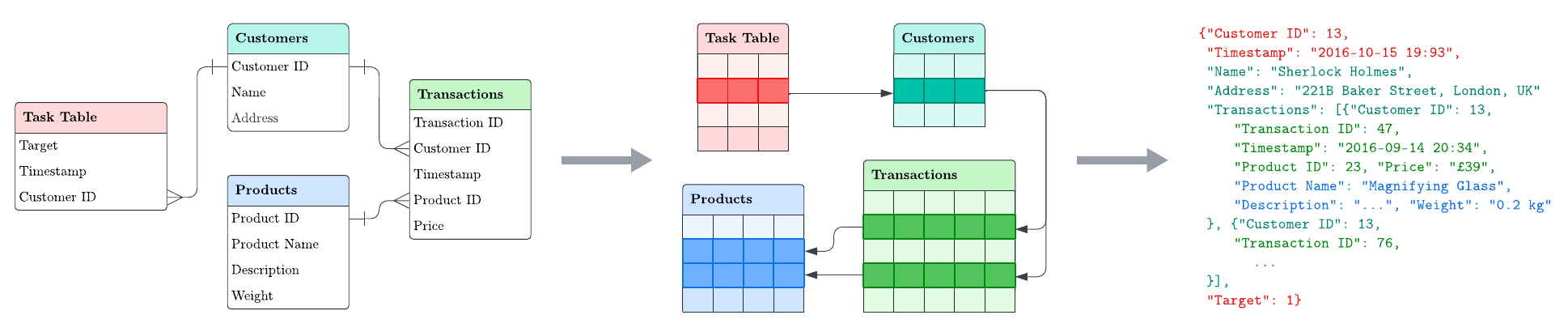}
\caption{Process of constructing a single example for LLM-based inference.}
\label{fig:example-text-gen}
\end{figure*}

Applying naive denormalization to the task table that follows only links in the direction from the foreign key to the primary key and appending columns from the linked table only partially solves the problem.
Much important information for the task is usually stored in the links in the opposite direction, from the primary key to the foreign key(s).
However, following these one-to-many links requires using an aggregation function, usually selected by human experts in the process of feature engineering.
\looseness=-1

To address the problem of feature engineering in relational databases, \citet{robinson2024relbench} proposed the use of graph neural networks. They first use deep tabular models that encode each row's attributes into initial entity embedding, %
mainly ResNet tabular model \citep{gorishniy2021revisiting} or textual embeddings based on word vectors such as GloVe~\citep{pennington2014glove} or contextual representations like those derived from BERT~\citep{devlin2019bert}.
These initial node embeddings are then fed into a %
graph neural network (GNN) \citep{gilmer2017neural, hamilton2017inductive} iteratively updating the node embeddings based on their neighbors.
In their work, they used two different architectures, the GraphSAGE model \citep{hamilton2017inductive} and ID-GNN \citep{you2021identity}.
Output node embeddings are fed into task-specific prediction heads and are learned end-to-end.
\citet{fey2023relational,robinson2024relbench} refer to this approach as Relational Deep Learning (RDL).

In this work, we investigated the feasibility of applying a much simpler approach to solving the task. We propose representing the prediction problem as a text document and allowing a pre-trained large language model to predict the output.
Our framework consists of two parts: 1) construction of the text document and 2) task-aware inference.

\subsection{Constructing documents for LLMs}
\label{sec:document-construction}

Our approach starts with the construction of a document for each sample $\bx$ in the test set. The document consists of three parts. First, we include the task context, that is, a short description of a relational database followed by a short description of the prediction task. We use the same descriptions as provided by \citet[Section 4 and Appendix A]{robinson2024relbench}.
The next part can consist of $\NInContext$, a specified number of in-context examples (including the predicted values) that are entities from the train table to give an additional demonstration of the task.
The in-context examples can be followed by the maximum $\NRelated$ latest related examples, which are entities from the train table linked to the same set of primary keys.
For every in-context example $v$, we ensure that $t_v < t_p$.
Finally, the entity for which the prediction needs to be made is added.

We apply a denormalization process to every entity added to the document, following links from a foreign key to a primary key.
We gather links related to all entities joined in the process of denormalization.
Following the links in the direction from a primary key to a foreign key, we select up to $\NNested$ entities from these tables.
We recursively apply the same procedure to them in breath-first order, up to the recursion depth $d$, but skip the tables already visited in previous denormalization steps.
Note that this process can be performed very efficiently on the fly on the database if hash indexes on all primary and foreign keys are constructed.
For binary classification, we use stratified sampling to obtain in-context examples to ensure both negative and positive samples.
To speed up the generation, we use the same set of in-context examples for every document for a given task.

As a result, we get an entity representation that includes nested entities from other tables. These entities are then serialized as JSONs. 
We choose this format, as it was demonstrated, that JSON performs well as the text representation of tabular data for LLMs \citet{singha2023tabular}.
Additionally, the JSON format allows for placing nested rows from other tables within the example.
This reduces the requirement for LLM to perform multi-hop inference, which often negatively impacts performance
\citep{wu2024mrke,Wu2024EvaluatingLI}.
The target variable is always listed as the last field in the serialized representation.
The process of generating serialized examples is demonstrated in \cref{fig:example-text-gen}.
In \cref{app:formal-algorithm}, we present the pseudocode of the entire procedure.
While the other text representations can be used instead of JSON, we do not investigate their impact on performance in this work.

\subsection{Metric-aware inference}

\begin{table*}[ht]

\small
\centering

\begin{tabular}{ll|rr|rrrr}
\toprule
Dataset & Task & LightGBM & RDL & Llama 3.2 1B & + MLP & Llama 3.2 3B & + MLP \\
\midrule
\multirow{2}{*}{rel-amazon} & user-churn & 52.22 & \textbf{70.42} & 60.56 & 66.56 & 62.55 & \textit{66.71} \\
 & item-churn & 62.54 & \textbf{82.81} & 71.96 & \textit{80.16} & 73.41 & 78.89 \\
\midrule
\multirow{2}{*}{rel-stack} & user-engagement & 63.39 & \textbf{90.59} & 81.01 & \textit{87.09} & 81.23 & 85.88 \\
 & user-badge & 63.43 & \textbf{88.86} & 71.13 & \textit{88.19} & 79.99 & 87.60 \\
\midrule
rel-trial & study-outcome & \textit{70.09} & 68.60 & 55.72 & 68.38 & 59.17 & \textbf{70.82} \\
\midrule
\multirow{2}{*}{rel-f1} & driver-dnf & 68.85 & 72.62 & 65.81 & 78.41 & \textit{80.03} & \textbf{82.33} \\
 & driver-top3 & 73.93 & 75.54 & \textit{88.47} & 87.36 & 87.11 & \textbf{89.70} \\
\midrule
rel-hm & user-churn & 55.21 & \textbf{69.88} & 64.34 & \textit{68.72} & 63.81 & 68.60 \\
\midrule
\multirow{2}{*}{rel-event} & user-repeat & 68.04 & \textbf{76.89} & 76.38 & \textit{76.72} & 70.11 & 73.88 \\
 & user-ignore & 79.93 & 81.62 & 78.55 & \textit{84.02} & 68.65 & \textbf{84.04} \\
\midrule
\multirow{2}{*}{rel-avito} & user-visits & 53.05 & \textbf{66.20} & 60.28 & \textit{64.98} & 53.36 & 64.24 \\
 & user-clicks & 53.60 & 65.90 & 61.32 & \textit{71.31} & 54.07 & \textbf{72.38} \\
\midrule
\multicolumn{2}{c|}{Average} & 63.69 & 75.83 & 69.63 & \textit{76.83} & 69.46 & \textbf{77.09} \\
\bottomrule
\end{tabular}

\caption{Comparison of LLMs' best-achieved results (AUROC, higher is better) with RDL and GBT models on entity classification tasks in RelBench. The best values are in
\textbf{bold}, second best in \textit{italic}.}
\label{tab:main-auroc}
\end{table*}

\begin{table*}[ht]

\small
\centering

\begin{tabular}{ll|rr|rr}
\toprule
Dataset & Task & LightGBM & RDL & Llama 3.2 1B + MLP & Llama 3.2 3B + MLP \\
\midrule
\multirow{2}{*}{rel-amazon} & user-ltv & 16.783 & \textbf{14.313} & 14.864 & \textit{14.789} \\
 & item-ltv & 60.569 & \textbf{50.053} & 52.682 & \textit{51.178} \\
\midrule
rel-stack & post-votes & \textit{0.068} & \textbf{0.065} & 0.090 & 0.085 \\
\midrule
\multirow{2}{*}{rel-trial} & study-adverse & \textbf{44.011} & \textit{44.473} & 51.845 & 48.383 \\
 & site-success & \textit{0.425} & \textbf{0.400} & 0.441 & 0.439 \\
\midrule
rel-f1 & driver-position & 4.170 & 4.022 & \textit{3.539} & \textbf{3.092} \\
\midrule
rel-hm & item-sales & 0.076 & \textbf{0.056} & \textit{0.057} & 0.064 \\
\midrule
rel-event & user-attendance & \textit{0.264} & \textbf{0.258} & 0.293 & 0.323 \\
\midrule
rel-avito & ad-ctr & \textit{0.041} & 0.041 & \textbf{0.036} & 0.041 \\
\midrule
\multicolumn{2}{c|}{Average} &  14.045 & \textbf{12.631} & 13.761 & \textit{13.155} \\
\bottomrule
\end{tabular}

\caption{Comparison of LLMs' best-achieved results (MAE, lower is better) with RDL and GBT models on entity regression tasks in RelBench. The best values are in
\textbf{bold}, second best in \textit{italic}.}
\label{tab:main-mae}

\end{table*}

We found that using a base version of LLM to simply fill out the tokens representing the target or asking for an instruct version for prediction
is not working well for such generated documents.
Therefore, following the work of \citet{lukasik2024metric}, we use base models and examine the probability distribution of possible subsequent tokens.
Under the assumption that a base model is trained only with the next-token prediction task using cross-entropy loss,
the minimizer under an unrestricted hypothesis class is the true conditional distribution $\mathbb{P}(y | \bx)$ \citep{gneiting2007strictly}.

In RelBench, entity classification tasks are evaluated using the AUROC metric.
It can be shown that the optimal decision rule for AUROC takes the form of any strictly monotone
transformation of $\mathbb{P}(y = 1 | \bx)$ \citep{clemenccon2008ranking, uematsu2014statistical}.
Because of that, we simply use the probability of the token representing a positive class (``1'').

For the regression tasks, RelBench uses the mean absolute error for evaluation, for which the optimal prediction is the median of $\mathbb{P}(y | \bx)$ \citep{bishop2006pattern}.
In the case of LLMs, the median can be calculated by sampling probabilities of values between minimum and maximum values seen in the train set.

\subsection{Applying simple prediction head}

We find that using metric-aware inference leads to good AUROC results in binary classification tasks. However, the median prediction approach performs poorly in regression tasks.
Because of that, to further test the strength of the representation produced by LLMs, instead of looking at the predicted distribution of tokens,
we propose training a small multilayered perception (MLP) head
using a small subset of training documents of size $\NTrain$ generated following the same process as outlined in \cref{sec:document-construction}.

\section{Experimental results}

In this section, we compare our method against the baselines reported in \citep{robinson2024relbench},
that is, with a simple baseline obtained using a LightGBM model~\citep{ke2017lightgbm} on naively normalized task tables and the RDL approach.
For each task, we generate documents for test-set examples using different combinations of
in-context examples $\NInContext \in \{0, 8, 16\}$, related examples $\NRelated \in \{0, 8, 16\}$, nested rows from linked tables $\NNested \in \{0, 4, 8\}$,
and depth of relation graph traversal $d \in \{0, 1\}$.
For our experiments, we use the two recently released Llama 3.2 models in sizes of 1B and 3B parameters \citep{dubey2024llama}, that support context-size up to 128k tokens. This is important as our method may result in documents with a large number of tokens. We report the average token count per document generated using different combinations of parameters in 
\cref{app:extended-results}. %

For the variant with trained MLP heads, we use either $\NTrain \in \{1\mathrm{e}{4}, 1\mathrm{e}{5}\}$ and MLP with a single hidden layer of size 10.
The examples from the available validation dataset were not included in the related examples $\NRelated$ and were not used for training the MLP.
More details can be found in \cref{app:experimental-details}.

We compare our approach with both the baselines on entity-wise classification and regression tasks from RelBench, for which we present the results in \cref{tab:main-auroc,tab:main-mae}.
In these tables, we report the result achieved by the best set of parameters selected for each task using a validation set. The detailed results for each parameter set are reported in \cref{app:extended-results}.
Given that some RelBench tasks have hundreds of thousands of test points, we limit our evaluation to reporting results on a uniform random sample of 10,000 test examples. 
Detailed statistics for the RelBench tasks are provided in \cref{app:relbench-stats}.

Metric-aware LLM inference consistently beats the LightGBM baseline on classification tasks but usually performs below RDL. %
However, by incorporating specialized MLP heads, our LLM-based approach achieves comparable or superior performance to RDL, 
notably, in many cases, with the MLP heads being trained only on a small subset of all available training data, as several RelBench datasets contain more than one million training examples. 
Since RDL requires end-to-end training, our approach potentially offers reduced resource requirements and may excel in low-data regimes, with specific advantages depending on the task and chosen LLM.

By checking the specific document parameters behind each result, we can observe that different information is important depending on the task. 
Some tasks benefit mainly from a large number of $\NRelated$, while others require information from linked tables (higher values of $\Depth$ and $\NNested$).
In almost all cases, LLMs perform poorly when tasked with zero-shot prediction, that is, the documents without any in-context/related example ($\NInContext = 0$, $\NNested = 0$, and $d = 0$), or when only simple in-context examples are provided ($\NInContext > 0$, $\NNested = 0$, and $d = 0$).
The only exception is the Rel-f1 dataset, which is a real database of results from Formula-1 races. In this case, LLMs exhibit strong performance even with just a few in-context examples, without additional information included.
We hypothesize that in this case, models seem to be mostly relying on their pre-existing factual knowledge of Formula-1, with the larger 3B model performing significantly better. Adding more information seems to only confuse models, resulting in degraded performance.

For other tasks, while the LLMs cannot leverage their factual knowledge, they effectively identify relationships between context information and the target variable $y$ when provided with relevant data. Though the 3B model generally outperforms its 1B counterpart, the performance gap is relatively modest, and all the results are close to those of the RDL approach.
This indicates that providing the context with important information is much more relevant than the parameter count.

\section{Conclusion}

In this preliminary investigation, we showed that LLMs can be used successfully for the prediction
tasks in the environment of relational databases without the need for additional fine-tuning by creating documents consisting of information from related entities.
We hypothesize that training models on tabular and relational data (similar to works of
\citet{tran2024tabularfm, li2024rllm}) may further enhance the performance of the proposed approach.
Additionally, we demonstrated that different tasks may require including different information in the context.
So far, we have selected the best combination of document generation parameters through a naive search using a validation dataset.
However, the size of documents quickly grows with the number of examples included and the depth of nesting entities.
Due to the limited context sizes of LLMs, a more intelligent selection of information (e.g., which columns to include and links to follow) may improve predictive performance as well as reduce computational cost.
Our approach also easily extends to data databases with datatypes that cannot be easily represented as text (e.g., images, audio) as multi-modal foundation models are recently gaining popularity.

\section*{Limitations}

We test our approach on a single, though arguably diverse, relational benchmark.
For specific databases and tasks, the input window required by the in-context learning we propose here might be too long for many popular LLMs.

\section*{Acknowledgments}

We thank Daniel Campos and Anupam Datta 
for their useful comments that helped us improve the readability of this work.

\bibliography{ms}

\newpage
\appendix
\onecolumn

\newcommand{\calV}{\mathcal{V}}

\section{RelBench datasets statistics}
\label{app:relbench-stats}

In \cref{tab:rel-datasets,tab:rel-tasks}, we report statistics of benchmarks used in this work.

\begin{table*}[ht]
\centering

\small
\begin{tabular}{l|l|r|r|r|r|r|r}
\toprule
Dataset        & Domain     & \#Tables & \#Rows      & \#Columns & Start date      & Validation date       & Test date      \\
\midrule
rel-amazon  & E-commerce & 3  & 15,000,713  & 15  & 2008-01-01 & 2015-10-01 & 2016-01-01 \\ 
rel-avito   & E-commerce & 8  & 20,679,117  & 42  & 2015-04-25 & 2015-05-08 & 2015-05-14 \\ 
rel-event   & Social     & 5  & 41,328,337  & 128 & 1912-01-01 & 2012-11-21 & 2012-11-29 \\ 
rel-f1      & Sports     & 9  & 74,063      & 67  & 1950-05-13 & 2005-01-01 & 2010-01-01 \\ 
rel-hm      & E-commerce & 3  & 16,664,809  & 37  & 2019-09-07 & 2020-09-07 & 2020-09-14 \\ 
rel-stack   & Social     & 7  & 4,247,264   & 52  & 2009-02-02 & 2020-10-01 & 2021-01-01 \\ 
rel-trial   & Medical    & 15 & 5,434,924   & 140 & 2000-01-01 & 2020-01-01 & 2021-01-01 \\ 
\bottomrule
\end{tabular}

\caption{Statistics of RelBench datasets. Datasets vary significantly in the number of tables, total number of rows, and number of columns. In this table, only counts of rows available for test inference are reported, i.e., rows up to the test time cutoff.}
\label{tab:rel-datasets}
\end{table*}

\begin{table*}[ht]
\centering
\small
\begin{tabular}{l|l|r|r|r|r|r}
\toprule
Dataset      & Task               & \multicolumn{3}{|c|}{\#Rows of task table} & \#Unique  & \%Train/Test \\ 
             &                    & Train   & Validation & Test   &  Entities          & Entity Overlap           \\ 
\midrule
rel-amazon   & user-churn         & 4,732,555 & 409,792 & 351,885 & 1,585,983 & 88.0 \\ 
             & item-churn         & 2,559,264 & 177,689 & 166,482 & 416,352 & 93.9 \\ 
             & user-ltv           & 4,732,555 & 409,792 & 351,885 & 1,585,983 & 88.0 \\ 
             & item-ltv           & 2,707,679 & 166,978 & 178,384 & 427,537 & 93.7 \\ 
\midrule
rel-avito    & ad-ctr             & 5,100 & 1,766 & 1,816 & 4,997 & 59.8 \\ 
             & user-clicks        & 59,454 & 21,183 & 47,996 & 66,449 & 44.3 \\ 
             & user-visits        & 86,619 & 29,979 & 36,129 & 13,405 & 64.6 \\ 
\midrule
rel-event    & user-attendance    & 19,261 & 2,014 & 2,006 & 9,694 & 14.6 \\ 
             & user-repeat        & 3,842 & 268 & 246 & 1,154 & 11.5 \\ 
             & user-ignore        & 19,239 & 4,185 & 4,010 & 9,979 & 21.1 \\ 
\midrule
rel-f1       & driver-dnf         & 11,411 & 566 & 702 & 821 & 50.0 \\ 
             & driver-top3        & 1,353 & 588 & 726 & 134 & 50.0 \\ 
             & driver-position    & 7,533 & 499 & 864 & 4,430 & 44.6 \\ 
\midrule
rel-hm       & user-churn         & 3,871,410 & 76,556 & 74,575 & 1,002,984 & 89.7 \\ 
             & item-sales         & 5,488,184 & 105,542 & 105,542 & 1,005,542 & 100.0 \\ 
\midrule
rel-stack    & user-engagement    & 1,360,850 & 85,838 & 88,137 & 88,137 & 97.4 \\ 
             & user-badge         & 3,386,276 & 247,398 & 255,360 & 255,360 & 98.9 \\ 
             & post-votes         & 2,453,921 & 156,216 & 160,903 & 160,903 & 97.1 \\ 
\midrule
rel-trial    & study-outcome      & 11,994 & 960 & 825 & 13,729 & 0.0 \\ 
             & study-adverse      & 43,335 & 3,596 & 3,098 & 50,029 & 0.0 \\ 
             & site-success       & 276,474 & 12,687 & 12,927 & 129,642 & 9.9 \\ 
\bottomrule
\end{tabular}
\caption{List of RelBench entity-wise prediction tasks with a number of train, validation, and test examples.}
\label{tab:rel-tasks}
\end{table*}

\section{Formal algorithm}
\label{app:formal-algorithm}

In \cref{alg:document-generation} we present the pseudocode of the procedure that constructs document representation $D_X$ of an example $X$, as described in \cref{sec:document-construction}. Here, we assume that examples $X$ behave as a dictionary containing a set of objects identifiable by keys. We denote a set of visited tables as $\calV$.

\begin{algorithm*}[ht!]
\caption{\textsc{Generate document}$(X, T_{\mathrm{train}}, \Tables, \Links, \NInContext, \NRelated, \Depth, \NNested)$}
\label{alg:document-generation}
\small 
\begin{algorithmic}[1]
\State \textbf{requires:} data and task descriptions, test entity $X$, train task table dataset $T_{\mathrm{train}} = [(X_i, y_i)]_{i=1}^{n_\mathrm{train}}$ , relation database $(\Tables, \Links)$, number of in-context examples $\NInContext$, number of related examples $\NRelated$, maximum depth of database traversal $\Depth$, maximum number of nested rows $\NNested$

\vspace{10pt}
\Procedure{Add related entities}{$X, \Tables, \Links, \NNested, \Depth_{\mathrm{cur}}, \Depth_{\max}, \calV$}
    \For{$(\TableFKey, \TablePKey) \in \Links: \mathrm{fkey} \in X'$}
        \State $X \gets X \cup \TablePKey$
        \State $\calV \gets \calV \cup \TablePKey$
    \EndFor
    \If{$\Depth = \Depth_{\max}$}
        \textbf{return} $X$
    \EndIf
    \For{$(\TableFKey, \TablePKey) \in \Links: \mathrm{pkey} \in X' $}
        \For{$i \in [\NNested]$}
            \State $X' \gets \arg \max_{(X'' \in \TableFKey: t_{X''} < t_{X} \wedge X'' \notin X)} t_{X''}$
            \State $X' \gets$ \textsc{Add related entities}$(X', \Tables, \Links, \NNested, \Depth_{\mathrm{cur}}, \Depth_{\max}, \calV)$ 
            \State $X \gets X \cup X'$
        \EndFor
        \State $\calV \gets \calV \cup \TableFKey$
    \EndFor
    \State \textbf{return} $X'$
\EndProcedure

\vspace{10pt}
\State $D_X \gets \text{data description} + 
\text{task description}$ 
\For{$i \in [\NInContext]$}
    \State $(X', y') \gets$ sample from $T_{\mathrm{train}}$ such that timestamp $t_{X'} < t_{X}$
    \State $X' \gets$ \textsc{Add related entities}$(X', \Tables, \Links, \NNested, 0, \Depth, \emptyset)$
    \State $D_X \gets D_X+$ \textsc{JSON}$(X', y')$
\EndFor
\For{$i \in [\NRelated]$}
    \State $(X', y') \gets \arg \max_{((X'', y'') \in T_{\mathrm{train}}: t_{X''} < t_{X} \wedge X'' \notin D_X)} t_{X''}$
    \State $X' \gets$ \textsc{Add related entities}$(X', \Tables, \Links, \NNested, 0, \Depth, \emptyset)$
    \State $D_X \gets D_X+$ \textsc{JSON}$(X', y')$
\EndFor
\State $D_X \gets D_X+$ \textsc{JSON}$(X)$
\State \textbf{return} $D_X$
\end{algorithmic}
\end{algorithm*}

\section{Experimental details}
\label{app:experimental-details}

All the experiments were conducted on a computational node with eight Nvidia H100 GPUs with 80GB of memory each. 
However, all the experiments can be reproduced using a single such GPU.
All the computation was performed using bfloat16 precision.

We use the same MLP architecture for all prediction tasks, that is, with the input of size of LLM's single token embedding (2048 for Llama 3.2 1B, and 3072 for Llama 3.2 3B),
one hidden layer of size 10 and an output. All MLP heads were trained using the Adam optimizer using an initial learning rate equal to $1\mathrm{e}{-4}$, 
which is linearly decreased for over 100 epochs, with the weight decay set to constant $1\mathrm{e}{-3}$.
The MLP is never trained on samples larger than 1e5 examples from the train set.
For validation purposes, we never use samples larger than 1e4 from the validation set.
If, at any point, the large number of tokens causes an out-of-memory error; we cut the number of $\NInContext$ and $\NRelated$ by half for the document causing the error.
We will publish the code for replicating all the results.

\section{Extended experimental results}
\label{app:extended-results}

In \cref{tab:ext-results-auroc,tab:ext-results-mae}, we present the detailed results for different combination of document parameters: $\NInContext \in \{0, 8, 16\}$, related examples $\NRelated \in \{0, 8, 16\}$, nested rows from linked tables $\NNested \in \{0, 4, 8\}$,
and depth of relation graph traversal $d \in \{0, 1\}$ and Llama 3.2 1B model. 
For each set of document generation parameters, we conducted a single run.
For the results of the Llama 3.2 3B model, we only tested selected combinations that turned out to be the most effective for the 1B variant.
In \cref{tab:token-stats}, we report the mean number of tokens resulting from the tokenization of documents generated with different combinations of parameters.

\begin{table*}[ht]

\resizebox{\linewidth}{!}{
\begin{tabular}{ll|rrrrrrrrrrrrr}
\toprule
\multicolumn{15}{c}{Llama 3.2 1B} \\
\midrule
Dataset & Task & \multicolumn{13}{|c}{Document parameters} \\
\midrule
                          &                      & $\NInContext = 0$ & $\NInContext = 8$ & $\NInContext = 8$ & $\NInContext = 0$ & $\NInContext = 0$ & $\NInContext = 8$ & $\NInContext = 8$ & $\NInContext = 16$ & $\NInContext = 16$ & $\NInContext = 0$ & $\NInContext = 0$ & $\NInContext = 16$ & $\NInContext = 16$ \\
                          &                      & $\NRelated = 0$   & $\NRelated = 0$   & $\NRelated = 0$   & $\NRelated = 8$   & $\NRelated = 8$   & $\NRelated = 8$   & $\NRelated = 8$   & $\NRelated = 0$    & $\NRelated = 0$    & $\NRelated = 16$  & $\NRelated = 16$  & $\NRelated = 16$   & $\NRelated = 16$  \\
                          &                      & $\Depth = 0$      & $\Depth = 0$      & $\Depth = 1$      & $\Depth = 0$      & $\Depth = 1$      & $\Depth = 0$      & $\Depth = 1$      & $\Depth = 0$       & $\Depth = 1$       & $\Depth = 0$      & $\Depth = 1$      & $\Depth = 0$       & $\Depth = 1$      \\
                          &                      & $\NRelated = 0$   & $\NRelated = 0$   & $\NRelated = 4$   & $\NRelated = 0$   & $\NRelated = 4$   & $\NRelated = 0$   & $\NRelated = 4$   & $\NRelated = 0$    & $\NRelated = 8$    & $\NRelated = 0$   & $\NRelated = 8$   & $\NRelated = 0$    & $\NRelated = 8$   \\

\midrule
\multirow{2}{*}{rel-amazon} & user-churn & 50.12 & 52.32 & 50.24 & 50.20 & 52.16 & 57.98 & 59.40 & 53.37 & 56.44 & 51.06 & 55.04 & \textbf{60.56} & \textit{60.37} \\
 & item-churn & 51.95 & 50.90 & 54.06 & 63.94 & 58.45 & \textbf{71.96} & 70.18 & 51.12 & 51.09 & 64.50 & 56.27 & \textit{70.33} & 70.09 \\
\midrule 
\multirow{2}{*}{rel-stack} & user-engagement & 54.98 & 63.54 & \textit{67.05} & \textbf{81.01} & - & - & - & - & - & - & - & - & - \\
 & user-badge & 51.44 & 52.45 & \textit{67.58} & \textbf{71.13} & - & - & - & - & - & - & - & - & - \\
\midrule 
rel-trial & study-outcome & 53.44 & 53.17 & 52.85 & \textit{53.75} & 50.16 & 53.17 & 52.85 & 51.82 & \textbf{55.72} & 53.75 & 51.21 & 51.82 & 52.32 \\
\midrule 
\multirow{2}{*}{rel-f1} & driver-dnf & 51.44 & 51.12 & 52.98 & \textit{64.95} & 57.99 & 59.24 & 53.29 & 55.84 & 52.24 & \textbf{65.81} & 60.22 & 62.04 & 63.55 \\
 & driver-top3 & 53.75 & \textbf{88.47} & 72.78 & 62.65 & 72.37 & 77.40 & 69.82 & \textit{78.65} & 71.80 & 66.01 & 68.93 & 57.79 & 50.56 \\
\midrule 
rel-hm & user-churn & 52.87 & 54.53 & 54.86 & 58.51 & 57.09 & \textit{62.12} & 52.85 & 54.07 & 55.36 & 59.99 & 57.61 & \textbf{64.34} & 58.97 \\
\midrule 
\multirow{2}{*}{rel-event} & user-repeat & 53.02 & 67.43 & 60.02 & 67.43 & 65.31 & \textbf{76.38} & 69.98 & 56.96 & 53.80 & 66.97 & 69.93 & 74.19 & \textit{74.58} \\
 & user-ignore & 53.68 & 66.46 & \textit{71.09} & 57.19 & 54.73 & 66.16 & \textbf{78.55} & 61.44 & 69.37 & 57.43 & 51.93 & 64.74 & 63.06 \\
\midrule 
\multirow{2}{*}{rel-avito} & user-visits & 50.54 & 50.42 & \textbf{60.28} & 50.45 & 60.09 & 50.43 & \textit{60.22} & 50.38 & 59.64 & 50.55 & 59.79 & 50.49 & 59.59 \\
 & user-clicks & 51.92 & 50.89 & \textit{60.34} & 52.19 & \textbf{61.32} & 51.75 & 59.44 & 51.90 & 58.33 & 51.35 & 58.84 & 52.30 & 59.12 \\
\bottomrule
\\
\toprule
\multicolumn{15}{c}{Llama 3.2 1B + MLP} \\
\midrule
Dataset & Task & \multicolumn{13}{|c}{Document parameters} \\
\midrule
                          &                      & $\NInContext = 0$ & $\NInContext = 8$ & $\NInContext = 8$ & $\NInContext = 0$ & $\NInContext = 0$ & $\NInContext = 8$ & $\NInContext = 8$ & $\NInContext = 16$ & $\NInContext = 16$ & $\NInContext = 0$ & $\NInContext = 0$ & $\NInContext = 16$ & $\NInContext = 16$ \\
                          &                      & $\NRelated = 0$   & $\NRelated = 0$   & $\NRelated = 0$   & $\NRelated = 8$   & $\NRelated = 8$   & $\NRelated = 8$   & $\NRelated = 8$   & $\NRelated = 0$    & $\NRelated = 0$    & $\NRelated = 16$  & $\NRelated = 16$  & $\NRelated = 16$   & $\NRelated = 16$  \\
                          &                      & $\Depth = 0$      & $\Depth = 0$      & $\Depth = 1$      & $\Depth = 0$      & $\Depth = 1$      & $\Depth = 0$      & $\Depth = 1$      & $\Depth = 0$       & $\Depth = 1$       & $\Depth = 0$      & $\Depth = 1$      & $\Depth = 0$       & $\Depth = 1$      \\
                          &                      & $\NRelated = 0$   & $\NRelated = 0$   & $\NRelated = 4$   & $\NRelated = 0$   & $\NRelated = 4$   & $\NRelated = 0$   & $\NRelated = 4$   & $\NRelated = 0$    & $\NRelated = 8$    & $\NRelated = 0$   & $\NRelated = 8$   & $\NRelated = 0$    & $\NRelated = 8$   \\

\midrule
\multirow{2}{*}{rel-amazon} & user-churn & 56.42 & 56.88 & 45.77 & \textit{66.46} & 65.28 & 65.85 & 59.67 & 56.75 & 51.21 & \textbf{66.56} & 65.71 & 66.43 & 66.40 \\
 & item-churn & 61.47 & 61.49 & 58.75 & 78.78 & 77.97 & 78.58 & 76.42 & 62.26 & 56.51 & 78.92 & \textbf{80.16} & \textit{78.93} & 75.62 \\
\midrule 
\multirow{2}{*}{rel-stack} & user-engagement & 56.05 & 65.44 & 70.17 & 85.90 & 83.95 & 85.95 & 82.97 & 66.91 & 82.96 & \textbf{87.09} & \textit{86.98} & 86.86 & 79.22 \\
 & user-badge & 59.57 & 61.86 & 85.45 & 80.87 & 87.07 & 80.79 & 87.19 & 62.27 & \textit{87.47} & 84.17 & \textbf{88.19} & 83.53 & 82.13 \\
\midrule 
rel-trial & study-outcome & 64.61 & \textit{68.16} & 66.47 & 65.55 & 64.49 & 67.58 & 66.47 & 67.91 & 66.86 & 64.76 & 63.10 & \textbf{68.38} & 67.18 \\
\midrule 
\multirow{2}{*}{rel-f1} & driver-dnf & 71.10 & \textit{77.14} & 51.78 & 70.66 & 65.89 & 74.13 & 48.88 & \textbf{78.41} & 54.60 & 70.52 & 67.13 & 72.69 & 49.55 \\
 & driver-top3 & 60.92 & \textbf{87.36} & 71.27 & 65.36 & 74.32 & 68.78 & 42.90 & \textit{83.46} & 40.41 & 65.43 & 63.35 & 69.65 & 52.43 \\
\midrule 
rel-hm & user-churn & 55.24 & 55.11 & 51.64 & 68.45 & 64.08 & 68.16 & 64.47 & 56.03 & 48.32 & \textbf{68.72} & 66.74 & \textit{68.62} & 63.33 \\
\midrule 
\multirow{2}{*}{rel-event} & user-repeat & 50.43 & 62.22 & 54.20 & \textbf{76.72} & 71.91 & 76.49 & 73.28 & 65.11 & 53.40 & 75.13 & \textit{76.56} & 74.62 & 50.58 \\
 & user-ignore & 66.10 & 68.77 & 64.16 & \textit{83.51} & 81.43 & \textbf{84.02} & 80.22 & 68.53 & 62.52 & 83.50 & 80.11 & 83.02 & 77.15 \\
\midrule 
\multirow{2}{*}{rel-avito} & user-visits & 57.18 & 54.51 & 62.84 & 57.23 & 63.39 & 57.28 & 63.27 & 54.69 & 62.79 & 56.18 & \textit{64.18} & 56.62 & \textbf{64.98} \\
 & user-clicks & 65.96 & 60.34 & 65.57 & 66.91 & 65.73 & 66.16 & 70.03 & 58.68 & 68.11 & 67.92 & \textit{70.07} & 67.34 & \textbf{71.31} \\
\bottomrule
\end{tabular}
}

\caption{Comparison of Llama 3.2 1B results (AUROC, higher is better) on documents generated using different parameters. 
The best values are in \textbf{bold}, second best in \textit{italic}.}
\label{tab:ext-results-auroc}
\end{table*}

\begin{table*}[ht]
\resizebox{\linewidth}{!}{
\begin{tabular}{ll|rrrrrrrrrrrrr}
\toprule
\multicolumn{15}{c}{Llama 3.2 1B + MLP} \\
\midrule
Dataset & Task & \multicolumn{13}{|c}{Document parameters} \\
\midrule
                          &                      & $\NInContext = 0$ & $\NInContext = 8$ & $\NInContext = 8$ & $\NInContext = 0$ & $\NInContext = 0$ & $\NInContext = 8$ & $\NInContext = 8$ & $\NInContext = 16$ & $\NInContext = 16$ & $\NInContext = 0$ & $\NInContext = 0$ & $\NInContext = 16$ & $\NInContext = 16$ \\
                          &                      & $\NRelated = 0$   & $\NRelated = 0$   & $\NRelated = 0$   & $\NRelated = 8$   & $\NRelated = 8$   & $\NRelated = 8$   & $\NRelated = 8$   & $\NRelated = 0$    & $\NRelated = 0$    & $\NRelated = 16$  & $\NRelated = 16$  & $\NRelated = 16$   & $\NRelated = 16$  \\
                          &                      & $\Depth = 0$      & $\Depth = 0$      & $\Depth = 1$      & $\Depth = 0$      & $\Depth = 1$      & $\Depth = 0$      & $\Depth = 1$      & $\Depth = 0$       & $\Depth = 1$       & $\Depth = 0$      & $\Depth = 1$      & $\Depth = 0$       & $\Depth = 1$      \\
                          &                      & $\NRelated = 0$   & $\NRelated = 0$   & $\NRelated = 4$   & $\NRelated = 0$   & $\NRelated = 4$   & $\NRelated = 0$   & $\NRelated = 4$   & $\NRelated = 0$    & $\NRelated = 8$    & $\NRelated = 0$   & $\NRelated = 8$   & $\NRelated = 0$    & $\NRelated = 8$   \\

\midrule
\multirow{2}{*}{rel-amazon} & user-ltv & 16.725 & 16.737 & 16.742 & \textbf{14.864} & 16.635 & \textit{15.019} & 16.613 & 16.713 & 16.743 & 15.025 & 16.254 & 15.022 & 16.138 \\
 & item-ltv & 61.905 & 61.705 & 62.095 & \textit{53.450} & 54.790 & 53.597 & 58.282 & 61.528 & 61.847 & 53.939 & 56.666 & \textbf{52.682} & 55.370 \\
\midrule 
rel-stack & post-votes & 0.098 & 0.095 & 0.093 & 0.099 & 0.092 & 0.093 & \textbf{0.090} & \textit{0.091} & 0.104 & 0.099 & 0.110 & 0.108 & 0.114 \\
\midrule 
\multirow{2}{*}{rel-trial} & study-adverse & 53.655 & 53.749 & 53.116 & \textbf{52.281} & \textit{53.066} & 54.406 & 53.848 & 53.168 & 53.425 & 53.838 & 53.569 & 54.884 & 53.324 \\
 & site-success & 0.453 & 0.450 & 0.450 & 0.444 & 0.446 & 0.449 & 0.446 & 0.445 & 0.447 & 0.446 & \textit{0.442} & \textbf{0.441} & 0.449 \\
\midrule 
rel-f1 & driver-position & 3.989 & \textbf{3.539} & 4.271 & 4.235 & 4.369 & 4.460 & 4.002 & \textit{3.902} & 4.400 & 4.400 & 4.379 & 3.948 & 4.360 \\
\midrule 
rel-hm & item-sales & 0.092 & 0.095 & 0.097 & 0.070 & 0.070 & 0.073 & \textbf{0.057} & 0.098 & 0.099 & 0.062 & \textit{0.062} & 0.064 & 0.082 \\
\midrule 
rel-event & user-attendance & 0.340 & 0.410 & 0.311 & 0.309 & 0.305 & \textbf{0.293} & 0.315 & 0.369 & 0.313 & 0.332 & \textit{0.304} & 0.334 & 0.317 \\
\midrule 
rel-avito & ad-ctr & 0.044 & 0.044 & 0.044 & 0.043 & 0.043 & 0.043 & \textbf{0.036} & 0.045 & 0.044 & 0.044 & \textit{0.043} & 0.043 & 0.043 \\
\bottomrule
\end{tabular}
}

\caption{Comparison of Llama 3.2 1B results (MAE, lower is better) on documents generated using different parameters. 
The best values are in \textbf{bold}, second best in \textit{italic}.}

\label{tab:ext-results-mae}
\end{table*}

\begin{table*}[ht]
\resizebox{\linewidth}{!}{
\begin{tabular}{ll|rrrrrrrrrrrrr}
\toprule
Dataset & Task & \multicolumn{13}{|c}{Document parameters} \\
\midrule
                          &                      & $\NInContext = 0$ & $\NInContext = 8$ & $\NInContext = 8$ & $\NInContext = 0$ & $\NInContext = 0$ & $\NInContext = 8$ & $\NInContext = 8$ & $\NInContext = 16$ & $\NInContext = 16$ & $\NInContext = 0$ & $\NInContext = 0$ & $\NInContext = 16$ & $\NInContext = 16$ \\
                          &                      & $\NRelated = 0$   & $\NRelated = 0$   & $\NRelated = 0$   & $\NRelated = 8$   & $\NRelated = 8$   & $\NRelated = 8$   & $\NRelated = 8$   & $\NRelated = 0$    & $\NRelated = 0$    & $\NRelated = 16$  & $\NRelated = 16$  & $\NRelated = 16$   & $\NRelated = 16$  \\
                          &                      & $\Depth = 0$      & $\Depth = 0$      & $\Depth = 1$      & $\Depth = 0$      & $\Depth = 1$      & $\Depth = 0$      & $\Depth = 1$      & $\Depth = 0$       & $\Depth = 1$       & $\Depth = 0$      & $\Depth = 1$      & $\Depth = 0$       & $\Depth = 1$      \\
                          &                      & $\NRelated = 0$   & $\NRelated = 0$   & $\NRelated = 4$   & $\NRelated = 0$   & $\NRelated = 4$   & $\NRelated = 0$   & $\NRelated = 4$   & $\NRelated = 0$    & $\NRelated = 8$    & $\NRelated = 0$   & $\NRelated = 8$   & $\NRelated = 0$    & $\NRelated = 8$   \\

\midrule
\multirow{8}{*}{rel-amazon} & user-churn & 151.7 & 492.7 & 2319.7 & 275.0 & 1039.2 & 616.0 & 2707.2 & 821.7 & 8688.5 & 286.8 & 1950.2 & 956.8 & 9662.2 \\
 &  & {\footnotesize $\pm$ 1.7} & {\footnotesize $\pm$ 1.7} & {\footnotesize $\pm$ 439.3} & {\footnotesize $\pm$ 103.9} & {\footnotesize $\pm$ 784.2} & {\footnotesize $\pm$ 103.9} & {\footnotesize $\pm$ 784.2} & {\footnotesize $\pm$ 1.7} & {\footnotesize $\pm$ 847.0} & {\footnotesize $\pm$ 135.5} & {\footnotesize $\pm$ 2108.4} & {\footnotesize $\pm$ 135.5} & {\footnotesize $\pm$ 2108.4} \\
 & item-churn & 341.3 & 2749.3 & 4263.6 & 1507.6 & 2409.3 & 3915.6 & 5736.3 & 4745.3 & 12896.9 & 1946.6 & 4409.6 & 6350.6 & 15825.6 \\
 &  & {\footnotesize $\pm$ 256.8} & {\footnotesize $\pm$ 256.8} & {\footnotesize $\pm$ 500.4} & {\footnotesize $\pm$ 1844.3} & {\footnotesize $\pm$ 2004.5} & {\footnotesize $\pm$ 1844.3} & {\footnotesize $\pm$ 2004.5} & {\footnotesize $\pm$ 256.8} & {\footnotesize $\pm$ 765.0} & {\footnotesize $\pm$ 2745.8} & {\footnotesize $\pm$ 3707.7} & {\footnotesize $\pm$ 2745.8} & {\footnotesize $\pm$ 3707.7} \\
 & user-ltv & 150.7 & 507.7 & 2174.4 & 280.3 & 1038.1 & 637.3 & 2566.1 & 865.7 & 8004.9 & 292.9 & 1963.4 & 1007.9 & 8993.4 \\
 &  & {\footnotesize $\pm$ 1.7} & {\footnotesize $\pm$ 1.7} & {\footnotesize $\pm$ 435.6} & {\footnotesize $\pm$ 108.8} & {\footnotesize $\pm$ 773.8} & {\footnotesize $\pm$ 108.8} & {\footnotesize $\pm$ 773.8} & {\footnotesize $\pm$ 1.7} & {\footnotesize $\pm$ 881.0} & {\footnotesize $\pm$ 142.6} & {\footnotesize $\pm$ 2165.7} & {\footnotesize $\pm$ 142.6} & {\footnotesize $\pm$ 2165.7} \\
 & item-ltv & 335.9 & 2671.9 & 3578.2 & 1512.2 & 2372.3 & 3848.2 & 5050.3 & 4195.9 & 8861.4 & 1973.2 & 4212.4 & 5833.2 & 11680.4 \\
 &  & {\footnotesize $\pm$ 230.6} & {\footnotesize $\pm$ 230.6} & {\footnotesize $\pm$ 511.9} & {\footnotesize $\pm$ 1580.8} & {\footnotesize $\pm$ 1836.2} & {\footnotesize $\pm$ 1580.8} & {\footnotesize $\pm$ 1836.2} & {\footnotesize $\pm$ 230.6} & {\footnotesize $\pm$ 811.2} & {\footnotesize $\pm$ 2325.9} & {\footnotesize $\pm$ 3452.7} & {\footnotesize $\pm$ 2325.9} & {\footnotesize $\pm$ 3452.7} \\
\midrule
\multirow{6}{*}{rel-stack} & user-engagement & 203.3 & 1017.3 & 9144.8 & 918.0 & 14351.9 & 1732.0 & 21384.6 & 2519.3 & 34108.3 & 1424.3 & 28134.9 & 3740.3 & 58659.4 \\
 &  & {\footnotesize $\pm$ 53.3} & {\footnotesize $\pm$ 53.3} & {\footnotesize $\pm$ 1735.9} & {\footnotesize $\pm$ 495.6} & {\footnotesize $\pm$ 12475.2} & {\footnotesize $\pm$ 495.6} & {\footnotesize $\pm$ 12408.9} & {\footnotesize $\pm$ 53.3} & {\footnotesize $\pm$ 2767.4} & {\footnotesize $\pm$ 949.6} & {\footnotesize $\pm$ 28355.3} & {\footnotesize $\pm$ 949.6} & {\footnotesize $\pm$ 25529.4} \\
 & user-badge & 201.1 & 1140.1 & 7222.5 & 907.6 & 5869.4 & 1846.6 & 12189.1 & 1893.1 & 31970.6 & 1356.5 & 11049.3 & 3048.5 & 41636.7 \\
 &  & {\footnotesize $\pm$ 49.3} & {\footnotesize $\pm$ 49.3} & {\footnotesize $\pm$ 1365.9} & {\footnotesize $\pm$ 507.0} & {\footnotesize $\pm$ 9637.1} & {\footnotesize $\pm$ 507.0} & {\footnotesize $\pm$ 9595.0} & {\footnotesize $\pm$ 49.3} & {\footnotesize $\pm$ 1997.3} & {\footnotesize $\pm$ 970.1} & {\footnotesize $\pm$ 20722.4} & {\footnotesize $\pm$ 970.1} & {\footnotesize $\pm$ 19269.3} \\
 & post-votes & 650.9 & 3860.9 & 9138.4 & 4352.3 & 8199.0 & 7561.4 & 16132.2 & 7020.9 & 29308.6 & 6902.2 & 26399.0 & 13269.1 & 52932.6 \\
 &  & {\footnotesize $\pm$ 650.7} & {\footnotesize $\pm$ 650.7} & {\footnotesize $\pm$ 1399.6} & {\footnotesize $\pm$ 4596.5} & {\footnotesize $\pm$ 9823.9} & {\footnotesize $\pm$ 4570.2} & {\footnotesize $\pm$ 9766.1} & {\footnotesize $\pm$ 650.7} & {\footnotesize $\pm$ 2328.7} & {\footnotesize $\pm$ 6755.0} & {\footnotesize $\pm$ 23700.3} & {\footnotesize $\pm$ 6697.7} & {\footnotesize $\pm$ 22108.6} \\
\midrule
\multirow{6}{*}{rel-trial} & study-outcome & 1571.3 & 12155.3 & 23127.0 & 1569.3 & 2644.0 & 12155.3 & 23127.0 & 23674.3 & 42720.1 & 1569.3 & 2746.1 & 23674.3 & 42720.1 \\
 &  & {\footnotesize $\pm$ 564.7} & {\footnotesize $\pm$ 564.7} & {\footnotesize $\pm$ 881.4} & {\footnotesize $\pm$ 564.7} & {\footnotesize $\pm$ 881.4} & {\footnotesize $\pm$ 564.7} & {\footnotesize $\pm$ 881.4} & {\footnotesize $\pm$ 564.7} & {\footnotesize $\pm$ 889.2} & {\footnotesize $\pm$ 564.7} & {\footnotesize $\pm$ 889.2} & {\footnotesize $\pm$ 564.7} & {\footnotesize $\pm$ 889.2} \\
 & study-adverse & 1605.4 & 12156.4 & 21552.0 & 1603.4 & 2668.0 & 12156.4 & 21552.0 & 20424.4 & 37546.3 & 1603.4 & 2739.3 & 20424.4 & 37546.3 \\
 &  & {\footnotesize $\pm$ 664.4} & {\footnotesize $\pm$ 664.4} & {\footnotesize $\pm$ 1001.7} & {\footnotesize $\pm$ 664.4} & {\footnotesize $\pm$ 1001.7} & {\footnotesize $\pm$ 664.4} & {\footnotesize $\pm$ 1001.7} & {\footnotesize $\pm$ 664.4} & {\footnotesize $\pm$ 1013.2} & {\footnotesize $\pm$ 664.4} & {\footnotesize $\pm$ 1013.2} & {\footnotesize $\pm$ 664.4} & {\footnotesize $\pm$ 1013.2} \\
 & site-success & 167.1 & 757.1 & 1367.6 & 258.6 & 442.7 & 848.6 & 1528.7 & 1399.1 & 3183.1 & 264.0 & 589.0 & 1496.0 & 3434.0 \\
 &  & {\footnotesize $\pm$ 6.8} & {\footnotesize $\pm$ 6.8} & {\footnotesize $\pm$ 56.5} & {\footnotesize $\pm$ 158.9} & {\footnotesize $\pm$ 341.8} & {\footnotesize $\pm$ 158.9} & {\footnotesize $\pm$ 341.8} & {\footnotesize $\pm$ 6.8} & {\footnotesize $\pm$ 125.4} & {\footnotesize $\pm$ 178.6} & {\footnotesize $\pm$ 597.7} & {\footnotesize $\pm$ 178.6} & {\footnotesize $\pm$ 597.7} \\
\midrule
\multirow{6}{*}{rel-f1} & driver-dnf & 211.1 & 901.1 & 1773.3 & 439.2 & 1192.9 & 1129.2 & 2058.9 & 1601.1 & 3757.4 & 655.2 & 2113.5 & 2045.2 & 4313.5 \\
 &  & {\footnotesize $\pm$ 3.5} & {\footnotesize $\pm$ 3.5} & {\footnotesize $\pm$ 434.9} & {\footnotesize $\pm$ 320.2} & {\footnotesize $\pm$ 720.4} & {\footnotesize $\pm$ 320.2} & {\footnotesize $\pm$ 720.4} & {\footnotesize $\pm$ 3.5} & {\footnotesize $\pm$ 881.3} & {\footnotesize $\pm$ 640.2} & {\footnotesize $\pm$ 1450.7} & {\footnotesize $\pm$ 640.2} & {\footnotesize $\pm$ 1450.7} \\
 & driver-top3 & 209.1 & 914.1 & 1788.8 & 435.0 & 1191.3 & 1140.0 & 2072.3 & 1609.1 & 4476.4 & 614.5 & 2069.1 & 2014.5 & 4985.1 \\
 &  & {\footnotesize $\pm$ 3.5} & {\footnotesize $\pm$ 3.5} & {\footnotesize $\pm$ 433.8} & {\footnotesize $\pm$ 316.6} & {\footnotesize $\pm$ 716.0} & {\footnotesize $\pm$ 316.6} & {\footnotesize $\pm$ 716.0} & {\footnotesize $\pm$ 3.5} & {\footnotesize $\pm$ 879.4} & {\footnotesize $\pm$ 595.6} & {\footnotesize $\pm$ 1397.9} & {\footnotesize $\pm$ 595.6} & {\footnotesize $\pm$ 1397.9} \\
 & driver-position & 194.2 & 878.2 & 1616.1 & 376.7 & 984.2 & 1060.7 & 1844.2 & 1564.2 & 3840.7 & 525.7 & 1689.0 & 1895.7 & 4255.0 \\
 &  & {\footnotesize $\pm$ 3.6} & {\footnotesize $\pm$ 3.6} & {\footnotesize $\pm$ 474.5} & {\footnotesize $\pm$ 305.4} & {\footnotesize $\pm$ 753.3} & {\footnotesize $\pm$ 305.4} & {\footnotesize $\pm$ 753.3} & {\footnotesize $\pm$ 3.6} & {\footnotesize $\pm$ 947.7} & {\footnotesize $\pm$ 570.8} & {\footnotesize $\pm$ 1462.2} & {\footnotesize $\pm$ 570.8} & {\footnotesize $\pm$ 1462.2} \\
\midrule
\multirow{4}{*}{rel-hm} & user-churn & 199.1 & 1096.1 & 1748.2 & 765.1 & 1042.9 & 1662.1 & 2405.9 & 2021.1 & 3097.5 & 965.6 & 1641.1 & 2787.6 & 4184.1 \\
 &  & {\footnotesize $\pm$ 4.5} & {\footnotesize $\pm$ 4.5} & {\footnotesize $\pm$ 55.5} & {\footnotesize $\pm$ 341.5} & {\footnotesize $\pm$ 399.2} & {\footnotesize $\pm$ 341.5} & {\footnotesize $\pm$ 399.2} & {\footnotesize $\pm$ 4.5} & {\footnotesize $\pm$ 112.3} & {\footnotesize $\pm$ 604.4} & {\footnotesize $\pm$ 810.1} & {\footnotesize $\pm$ 604.4} & {\footnotesize $\pm$ 810.1} \\
 & item-sales & 367.4 & 2731.4 & 2957.6 & 2668.0 & 3439.8 & 5032.0 & 5909.8 & 4984.4 & 5874.5 & 4968.8 & 7257.0 & 9585.8 & 12546.0 \\
 &  & {\footnotesize $\pm$ 16.1} & {\footnotesize $\pm$ 16.1} & {\footnotesize $\pm$ 95.2} & {\footnotesize $\pm$ 145.3} & {\footnotesize $\pm$ 768.3} & {\footnotesize $\pm$ 145.3} & {\footnotesize $\pm$ 768.3} & {\footnotesize $\pm$ 16.1} & {\footnotesize $\pm$ 186.6} & {\footnotesize $\pm$ 274.5} & {\footnotesize $\pm$ 2589.6} & {\footnotesize $\pm$ 274.5} & {\footnotesize $\pm$ 2589.6} \\
\midrule
\multirow{6}{*}{rel-event} & user-repeat & 229.9 & 997.9 & 6334.6 & 503.1 & 2772.6 & 1271.1 & 8233.6 & 1747.9 & 14789.1 & 572.0 & 4972.1 & 2090.0 & 18558.1 \\
 &  & {\footnotesize $\pm$ 2.6} & {\footnotesize $\pm$ 2.6} & {\footnotesize $\pm$ 820.0} & {\footnotesize $\pm$ 277.0} & {\footnotesize $\pm$ 5357.0} & {\footnotesize $\pm$ 277.0} & {\footnotesize $\pm$ 5357.0} & {\footnotesize $\pm$ 2.6} & {\footnotesize $\pm$ 1331.1} & {\footnotesize $\pm$ 423.7} & {\footnotesize $\pm$ 12405.2} & {\footnotesize $\pm$ 423.7} & {\footnotesize $\pm$ 12405.2} \\
 & user-ignore & 217.2 & 992.2 & 3809.3 & 453.4 & 1653.8 & 1228.4 & 4911.8 & 1778.2 & 10950.1 & 516.2 & 2813.0 & 2077.2 & 13039.0 \\
 &  & {\footnotesize $\pm$ 2.7} & {\footnotesize $\pm$ 2.7} & {\footnotesize $\pm$ 514.1} & {\footnotesize $\pm$ 261.5} & {\footnotesize $\pm$ 3538.9} & {\footnotesize $\pm$ 261.5} & {\footnotesize $\pm$ 3538.9} & {\footnotesize $\pm$ 2.7} & {\footnotesize $\pm$ 792.2} & {\footnotesize $\pm$ 415.0} & {\footnotesize $\pm$ 8078.2} & {\footnotesize $\pm$ 415.0} & {\footnotesize $\pm$ 8078.2} \\
 & user-attendance & 207.9 & 994.9 & 2646.0 & 448.3 & 1707.0 & 1235.3 & 3780.0 & 1663.9 & 15828.4 & 511.5 & 2897.5 & 1967.5 & 17979.5 \\
 &  & {\footnotesize $\pm$ 2.6} & {\footnotesize $\pm$ 2.6} & {\footnotesize $\pm$ 544.5} & {\footnotesize $\pm$ 263.9} & {\footnotesize $\pm$ 3635.8} & {\footnotesize $\pm$ 263.9} & {\footnotesize $\pm$ 3635.8} & {\footnotesize $\pm$ 2.6} & {\footnotesize $\pm$ 822.4} & {\footnotesize $\pm$ 420.3} & {\footnotesize $\pm$ 8339.3} & {\footnotesize $\pm$ 420.3} & {\footnotesize $\pm$ 8339.3} \\
\midrule
\multirow{6}{*}{rel-avito} & user-visits & 197.0 & 197.0 & 558.0 & 197.0 & 558.0 & 197.0 & 558.0 & 197.0 & 826.1 & 197.0 & 826.1 & 197.0 & 826.1 \\
 &  & {\footnotesize $\pm$ 0.2} & {\footnotesize $\pm$ 0.2} & {\footnotesize $\pm$ 268.5} & {\footnotesize $\pm$ 0.2} & {\footnotesize $\pm$ 268.5} & {\footnotesize $\pm$ 0.2} & {\footnotesize $\pm$ 268.5} & {\footnotesize $\pm$ 0.2} & {\footnotesize $\pm$ 508.2} & {\footnotesize $\pm$ 0.2} & {\footnotesize $\pm$ 508.2} & {\footnotesize $\pm$ 0.2} & {\footnotesize $\pm$ 508.2} \\
 & user-clicks & 197.0 & 197.0 & 501.2 & 197.0 & 501.2 & 197.0 & 501.2 & 197.0 & 718.7 & 197.0 & 718.7 & 197.0 & 718.7 \\
 &  & {\footnotesize $\pm$ 0.2} & {\footnotesize $\pm$ 0.2} & {\footnotesize $\pm$ 269.8} & {\footnotesize $\pm$ 0.2} & {\footnotesize $\pm$ 269.8} & {\footnotesize $\pm$ 0.2} & {\footnotesize $\pm$ 269.8} & {\footnotesize $\pm$ 0.2} & {\footnotesize $\pm$ 505.4} & {\footnotesize $\pm$ 0.2} & {\footnotesize $\pm$ 505.4} & {\footnotesize $\pm$ 0.2} & {\footnotesize $\pm$ 505.4} \\
 & ad-ctr & 291.0 & 291.0 & 597.1 & 291.0 & 597.1 & 291.0 & 597.1 & 291.0 & 857.9 & 291.0 & 857.9 & 291.0 & 857.9 \\
 &  & {\footnotesize $\pm$ 30.3} & {\footnotesize $\pm$ 30.3} & {\footnotesize $\pm$ 71.4} & {\footnotesize $\pm$ 30.3} & {\footnotesize $\pm$ 71.4} & {\footnotesize $\pm$ 30.3} & {\footnotesize $\pm$ 71.4} & {\footnotesize $\pm$ 30.3} & {\footnotesize $\pm$ 138.8} & {\footnotesize $\pm$ 30.3} & {\footnotesize $\pm$ 138.8} & {\footnotesize $\pm$ 30.3} & {\footnotesize $\pm$ 138.8} \\
\bottomrule
\end{tabular}
}

\caption{Average token counts and standard deviations for documents generated with different parameters and tokenized using the Llama 3.2 tokenizer.}

\label{tab:token-stats}
\end{table*}

\end{document}